%% file: acl2023.tex
\pdfoutput=1

\documentclass[11pt]{article}

\usepackage{ACL2023}

\usepackage{times}
\usepackage{latexsym}

\input{math_commands.tex}

\usepackage{hyperref}
\usepackage{url}

\usepackage{booktabs}
\usepackage{multirow}
\usepackage{multicol}
\usepackage{amsmath}

\usepackage{microtype}
\usepackage{graphicx}
\usepackage{subfigure}
\usepackage{booktabs} 

\usepackage{bbm}
\usepackage{nicefrac}
\usepackage{color}
\definecolor{tab_blue}{rgb}{0.121,0.466,0.705}
\definecolor{tab_orange}{rgb}{1.000,0.498,0.055}
\definecolor{tab_green}{rgb}{0.172,0.627,0.172}

\usepackage{subfiles}

\usepackage[T1]{fontenc}

\usepackage[utf8]{inputenc}

\usepackage{microtype}

\usepackage{inconsolata}

\title{Downstream Datasets Make Surprisingly Good Pretraining Corpora}

\author{Kundan Krishna, Saurabh Garg, Jeffrey P. Bigham, Zachary C. Lipton \\
Carnegie Mellon University \\
  \texttt{\{kundank,sgarg2,jbigham,zlipton\}@andrew.cmu.edu} \\}

\begin{document}

\maketitle

\begin{abstract}
\input{sections/00_abstract}
\end{abstract}

\input{sections/10_intro}
\input{sections/20_relatedwork}

\input{sections/30_method}

\input{sections/40_self_pretraining}

\input{sections/50_crosstask_perf}

\input{sections/60_ensembling}
\input{sections/80_generality}

\input{sections/85_moretasks.tex}

\input{sections/90_conclusion}

\input{sections/95_limitations}

\section{Acknowledgements}
This work was funded by UPMC and Abridge AI Inc. We also gratefully acknowledge support from Amazon
AI, the PwC Center, the Software Engineering Institute, and NSF (Award no. 2211955) for the compute
resources used in this project.
SG acknowledges Amazon Graduate Fellowship and  JP Morgan AI Ph.D. Fellowship for their support. 

\bibliography{custom}
\bibliographystyle{acl_natbib}

\newpage

\appendix
\section{Appendix}
\input{sections/appendix}

\end{document}

%% file: math_commands.tex
\usepackage{amsmath,amsfonts,bm}

\def\eqref#1{equation~\ref{#1}}

\def\1{\bm{1}}

\DeclareMathAlphabet{\mathsfit}{\encodingdefault}{\sfdefault}{m}{sl}
\SetMathAlphabet{\mathsfit}{bold}{\encodingdefault}{\sfdefault}{bx}{n}



%% file: sections/00_abstract.tex
For most natural language processing tasks,
the dominant practice is to finetune 
large pretrained transformer models (e.g., BERT) 
using smaller downstream datasets.
Despite the success of this approach,
it remains unclear to what extent these gains
are attributable to the massive background corpora
employed for pretraining versus 
to the pretraining objectives themselves.
This paper introduces a large-scale study of \emph{self-pretraining},
where the same (downstream) training data
is used for both pretraining and finetuning.
In experiments addressing both ELECTRA and RoBERTa models
and 10 distinct downstream classification datasets, 
we observe that self-pretraining 
rivals
standard pretraining 
on the BookWiki corpus 
(despite using around $10\times$--$500\times$ less data), 
outperforming the latter on $7$ and $5$ datasets, respectively.
Surprisingly, these task-specific pretrained models
often perform well on other tasks,
including the GLUE benchmark.
Self-pretraining 
also provides benefits on structured output prediction tasks
such as question answering and commonsense inference,
often providing more than 50\% improvements 
compared to standard pretraining. 
Our results hint that often
performance gains attributable to pretraining 
are driven primarily by the pretraining objective itself 
and are not always attributable 
to the use of external pretraining data in massive amounts.
These findings are especially relevant 
in light of concerns about
intellectual property 
and offensive content in web-scale pretraining data.\footnote{Pretrained models can be downloaded from \url{https://github.com/acmi-lab/self-pretrain}}

%% file: sections/10_intro.tex
\section{Introduction}

For training predictive models operating on natural language data,
the current best practice is to \emph{pretrain} models
on large unlabeled \emph{upstream} corpora
to optimize self-supervised objectives, 
for example, masked language modeling (MLM);
the resulting weights are then used to initialize
models that are subsequently trained (\emph{finetuned})
on the labeled \emph{downstream} data available for the task at hand.
Large-scale pretrained models typically provide 
significant performance boosts when compared to models 
trained directly on the downstream task 
(with random initializations)
\citep{peters-etal-2018-deep, devlin2019bert, chiang2020pre, krishna2021does}.
Upstream corpora tend to be significantly larger than the downstream corpora 
and the success of this approach is often attributed 
to its ability to leverage these massive upstream corpora~\citep{liu2019roberta, yang2019xlnet}. 
For example, the seminal BERT model~\citep{devlin2019bert}
was pretrained using the BookWiki corpus which is 
a combination of English Wikipedia
and BooksCorpus~\citep{zhu2015aligning}, totaling 13GB of plain text.
Subsequent models have moved on to web-scale data.
For example, XLNet~\citep{yang2019xlnet},
RoBERTa~\citep{liu2019roberta},
and T5~\citep{raffel2020exploring}), 
were trained on 158GB, 160GB and 750GB of data, respectively.

\begin{figure*}
    \centering
    \includegraphics[width=0.9\textwidth]{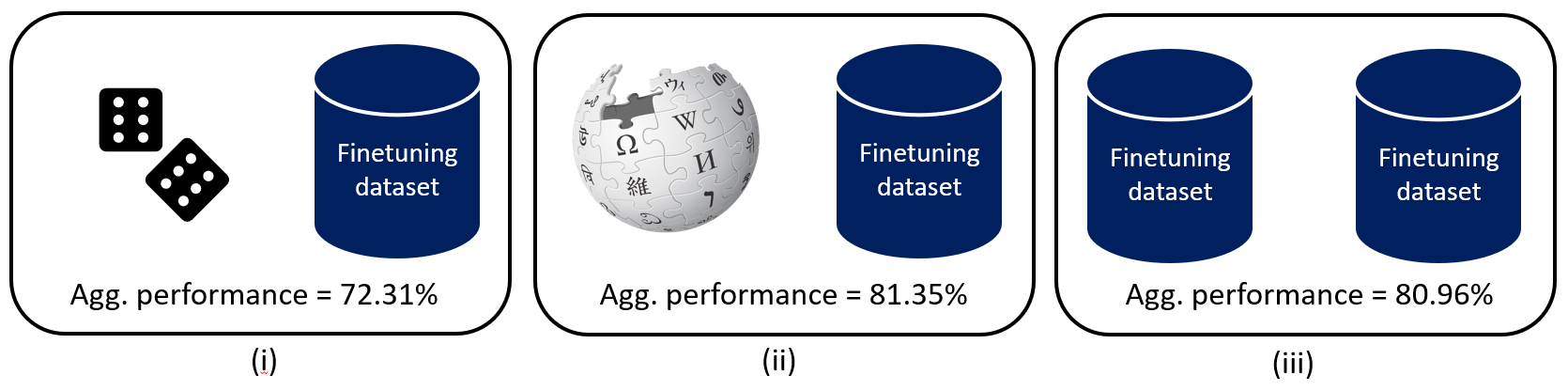}
    \caption{Aggregate performance of an ELECTRA model across 10 finetuning datasets when it is (i) randomly initialized (ii) pretrained on upstream corpus (BookWiki) (iii) pretrained on the finetuning dataset itself}
    \label{fig:figure1}
\end{figure*}

As upstream corpus size and downstream performance have gone up,
popular attempts at explaining these gains
have focused on themes of ``knowledge transfer" from the upstream corpus, 
attributing them to shared linguistic structure, semantics~\citep{lina2019open,tenney2019bert}, 
and facts about the world~\citep{petroni2019language}.
However, since the introduction of large-scale pretraining corpora 
occurred together with the invention of self-supervised pretraining objectives 
(e.g. masked language modeling~\citep{devlin2019bert}
and replaced token detection~\citep{clark2019electra}),
it remains unclear to what extent 
large-scale corpora are integral
to these leaps in performance.
For several tasks, especially summarization, 
recent works achieved 
surprising performance gains in settings 
where the upstream corpus is created synthetically 
with 
arbitrary symbols,
but the pretraining objective is designed
to capture some of the structure of the task
\citep{krishna2021does,wu2022insights}. 

In this work, we ask just how much of pretraining's benefits
could be realized in the absence of upstream corpora
by pretraining directly on the downstream corpora 
(with the same self-supervised objectives).
We find that this approach, which we call \emph{self-pretraining},
often rivals the performance boosts conferred 
by \emph{off-the-shelf} models pretrained
on large upstream corpora (Figure~\ref{fig:figure1}),
even outperforming them on 7 out of 10 datasets.
Prior research has shown that \emph{additional} 
self-supervised pretraining of off-the-shelf models 
using the downstream data can give further gains~\citep{gururangan2020don}.
Our study goes further, showing that 
even when starting from random initializations, 
and without using any external data beyond 
the downstream data itself,
self-pretraining can rival standard practices. 
Since self-pretraining requires the same data 
that must already be available for 
finetuning, 
the benefits of pretraining in this case 
cannot be attributed to \emph{transfer} of knowledge
from the upstream corpus.
Instead, these benefits can only be attributed 
to the pretraining objective,
which is possibly able to learn some inductive biases
better than the finetuning objective (e.g. linguistic knowledge~\cite{tenney2019bert}), or perhaps 
simply initializes network parameters such that their 
statistics lead to better optimization during finetuning~\citep{wu2022insights}.
While similar observations 
were made
in the computer vision community~\citep{el2021large},
we argue that it is especially important to establish
these phenomena in the language domain,
for which building on self-supervised pretrained models
is now a ubiquitous practice.

To understand differences in predictions 
with different pretraining strategies 
(i.e., between  self-pretrained and off-the-shelf models), 
we analyse the errors made by these models on the same downstream data (Sec.~\ref{sec:disagree}). 
Despite similar performance of these models,
we find that self-pretrained and off-the-shelf models
make significantly 
less correlated
errors
when compared to two independently finetuned models pretrained 
with either strategy.

We find that models pretrained on one downstream dataset
often perform surprisingly well
when finetuned to other downstream datasets (Sec.~\ref{sec:cross_finetuning}), including the GLUE benchmark.
Even though the downstream datasets in our study 
come from a wide variety of domains (e.g., news, online forums, tweets), 
we find that pretraining 
on any of these downstream datasets 
delivers significant performance gains on most datasets
(greater than half of off-the-shelf model's gains in $88\%$ of cases) 
irrespective of domain.
However, the best performance on a downstream dataset 
is usually achieved by the model pretrained on that dataset itself.

In addition to classification tasks, we also experiment with tasks
such as span-based question answering,
named entity recognition, and grounded commonsense inference (Sec. \ref{sec:SP_task}).
Self-pretraining delivers around 40-80\% of the performance boost compared to
models pretrained on the BookWiki corpus across ELECTRA and RoBERTa models.
Hence, self-pretraining can perform better than fine-tuning
randomly initialized models even for tasks that require 
prediction of more complex structured output than a single  label,
and  for tasks whose solution relies on commonsense knowledge.

Our contributions can be summarized as follows:
\begin{itemize}
\setlength{\itemsep}{0pt}
    \setlength{\parskip}{0pt}
    \setlength{\parsep}{0pt}  

    \item Comparison of self-pretrained and off-the-shelf pretrained models (both with ELECTRA and RoBERTa architectures) across 10 downstream classification tasks. 
    
    \item Analysis of out-of-distribution performance 
    of models pretrained on one downstream dataset 
    and finetuned on other downstream datasets, 
    including the GLUE benchmark.
    \item Demonstration of self-pretraining's efficacy on more complex tasks than classification
    such as tasks requiring  structured output prediction or  commonsense reasoning.
\end{itemize}

%% file: sections/20_relatedwork.tex
\section{Related Work}

\textbf{Self-pretraining in Computer Vision {} {}}
Most relevant to our work, 
recent/concurrent works in computer vision
explore self-pretraining~\citep{he2022masked, el2021large}.
In a contemporary work, \citet{he2022masked} showed 
that pretraining with a Masked AutoEncoder (MAE) objective 
(analogue of MLM objective for images) 
boosts the performance 
of ViT models on the Imagenet-1K dataset. 
\citet{el2021large}
showed that pretraining solely on 
downstream datasets 
for object detection and segmentation tasks reaches 
the performance of Imagenet-pretrained models.
Our work establishes that a similar phenomenon is observed
for NLP tasks too across a wide range of datasets.

\noindent \textbf{Pretraining on Downstream Data in NLP {} {}} 
\emph{Task-Adaptive PreTraining} 
(TAPT~\citep{gururangan2020don})
consists of taking off-the-shelf 
pretrained models like BERT and RoBERTa
and engaging in further pretraining 
on the downstream datasets 
before finetuning them to the task at hand. 
TAPT has been shown to improve performance 
of off-the-shelf models in a variety 
of works~\citep{logeswaran2019zero,han2019unsupervised,chakrabarty2019imho}.
Another way in which downstream data has been used is for retrieval to create a small pretraining corpus for efficient pretraining \citep{yao2022nlp}.
By contrast, our work pretrains models 
\emph{only} on the downstream dataset, 
enabling a head-to-head comparison
between the performance of off-the-shelf
and self-pretrained models, 
and (in some situations) challenging 
the necessity of upstream corpora altogether.

\noindent \textbf{Claims about Knowledge transfer {} {}}
Many works claim that pretraining
extracts generally useful \emph{knowledge} 
from the upstream corpus such as linguistic patterns~\citep{lina2019open,tenney2019bert,manning2020emergent} and facts~\citep{petroni2019language},
and that this accounts for the performance gains 
that they enjoy on downstream tasks.
Several works, e.g., in the \emph{probing} literature \citep{tenney2019bert, manning2020emergent, petroni2019language},
demonstrate that from the internal representations of a model,
it is easy (e.g., via linear models) 
to predict certain linguistic features or real-world facts.
However, these studies do not clarify the mechanism
by which these observations relate
to performance gains on downstream tasks.
\citet{tenney2019bert} recognizes this limitation, stating 
\emph{``the observation of a (linguistic) pattern 
does not tell us how it is used''}.
Our work suggests that to the extent that such knowledge extraction
plays a role in pretraining's benefits,
sufficient knowledge is often present in the downstream dataset
and need not be \emph{transferred} from huge upstream corpora.

\noindent \textbf{Challenges to the Knowledge Transfer Narrative {} {}}
Multiple previous works have questioned 
whether knowledge transfer can fully account
for the efficacy of pretraining.
Improvements in performance on downstream NLP tasks 
have resulted from pretraining on other modalities 
like music and code \citep{papadimitriou2020learning},
sequences of meaningless symbols ~\citep{chiang2020pre, krishna2021does, wu2022insights}, 
and language denatured via shuffling of words \citep{sinha2021masked}.
On the other hand, models pretrained on language 
have shown improved performance on tasks 
dealing with other modalities 
such as image classification~\cite{lu2021pretrained} 
and reinforcement learning for games~\cite{reid2022can}.
By contrast, we show that without surplus upstream data of any modality,
self-pretraining alone can often perform comparably 
or even better than standard pretraining with a large upstream corpus. 
In a similar vein with these papers, our work suggests 
that a large portion of pretraining's success
may come from alternative, unexplored mechanisms 
which have more to do with the pretraining objective 
than knowledge transfer from upstream corpora.

%% file: sections/30_method.tex
\section{Experimental Setup} 

\begin{table*}[t]
    \centering
    \footnotesize
    \renewcommand{\arraystretch}{1.2}
\begin{tabular}{lrrll}
\toprule
             \textbf{Dataset} &  \textbf{Size (MB)} &  \textbf{Classes} &                  \textbf{Domain} &                         \textbf{Task} \\
\midrule
              AGNews~\citep{Zhang2015CharacterlevelCN} &    27 &        4 &                    News &         topic classification \\
                 QQP~\citep{wang2018glue} &    43 &        2 &  Online forum questions &         paraphrase detection \\
     Jigsaw Toxicity~\citep{jigsaw_tox} &    59 &        6 &      Wikipedia comments &           toxicity detection \\
                MNLI~\citep{williams2018broad} &    65 &        3 &              Diverse &   natural language inference \\
        Sentiment140~\citep{go2009twitter} &   114 &        5 &                  Tweets &     sentiment classification \\
                PAWS~\citep{zhang2019paws} &   139 &        2 &               Wikipedia &         paraphrase detection \\
           DBPedia14~\citep{Zhang2015CharacterlevelCN} &   151 &       14 &               Wikipedia &         topic classification \\
           Discovery~\citep{sileo-etal-2019-mining} &   293 &      174 &               Web crawl &  discourse marker prediction \\
  Yahoo Answertopics~\citep{Zhang2015CharacterlevelCN} &   461 &       10 &    Online forum answers &         topic classification \\
     Amazon Polarity~\citep{Zhang2015CharacterlevelCN} &  1427 &        2 &         Product reviews &     sentiment classification \\
\bottomrule
\end{tabular}
    \caption{The suite of downstream datasets used in this work along with their training set sizes}
    \label{tab:dataset_details}
\end{table*}

Our experiments center around the ELECTRA model~\citep{clark2019electra} and the RoBERTa-base model~\citep{liu2019roberta}. 
On the broadest set of experiments, 
for which we can only afford to train one model, we employ ELECTRA because it performs better than RoBERTa given comparable compute budgets~\citep{clark2019electra}.
In particular, we use the small variant of ELECTRA (14 million parameters), which performs similarly to BERT-base on GLUE (difference of $\approx$2 points)
while training much faster~\citep{clark2019electra}.
However, we replicate many of these results on the larger RoBERTa-base model revealing similar results and thus establishing the generality of our findings.

During pretraining, a text sequence is fed into the model with some tokens masked out. 
While MLM-only models like RoBERTa
only have a \emph{generator} network
that predicts the content of the masked tokens, 
ELECTRA has an additional discriminator module
that predicts if those predictions were correct.
Both the generator and the discriminator networks' parameters 
are updated simultaneously during pretraining.
After pretraining, the generator is discarded 
and the discriminator is used as an encoder 
for finetuning on downstream tasks.

We experimented with 10 different downstream datasets 
(Table~\ref{tab:dataset_details}). 
We chose these datasets in our testbed to span different dataset sizes 
ranging from 27 megabytes  to about 1.4 gigabytes of text in the training split. 
These datasets are for different tasks such as topic classification, 
sentiment classification, natural language inference etc., 
and are created using data sourced from diverse domains. 
Most of them are multi-class classification tasks 
except Jigsaw Toxicity which is a multi-label classification task, 
and Sentiment140 which is modeled as a regression task.
For finetuning a pretrained model on any dataset, 
we passed the input through the model, 
took the vector representation of the CLS token in the final layer, 
and passed it through a classification head 
with one hidden layer to get the output.

%% file: sections/40_self_pretraining.tex
\section{Self-pretraining Performance}

In our first set of experiments, 
we compare self-pretraining's performance 
with other pretraining techniques.  
For each dataset, we pretrain an ELECTRA model 
on text from its training split and then finetune it 
on the same training data using the associated labels.
To create a pretraining corpus from a downstream dataset, 
we concatenate the input text from each of the examples, 
assembling them in random order.
We evaluate the performance of each finetuned model 
on the corresponding dataset's test split. For 
QQP and MNLI we just use the validation split 
because test set labels are private.
For all datasets, we evaluate performance by accuracy,
except for Sentiment140 and Jigsaw Toxicity,
for which we use Pearson correlation 
and micro-averaged AUC scores, respectively
(these are not multi-class classification problems).

Notably, all self-pretrained models deliver significant performance boosts 
on their respective datasets (Table~\ref{tab:self_pretrain}), 
and over half of them perform even better 
than the off-the-shelf model.
We measured a model's \emph{benefit} as the increase in performance metric that it achieves
over a randomly initialized model,
divided by the increase in performance metric achieved by the off-the-shelf ELECTRA model
against the same baseline.
The average benefit of self-pretraining across all datasets is $103.70\%$.
We do not see a clear correlation 
between the size of the dataset 
and the performance of self-pretraining. 
For example, the highest benefit of $131.33\%$ 
is achieved for the smallest dataset (AGNews),
which is merely 27MB in size,
while the minimum benefit is achieved on the Discovery dataset, 
which is the third largest dataset measuring 293MB.
For each downstream dataset, we also pretrain a model on a randomly sampled subset of Wikipedia of the same size as the dataset's training corpus, and finetune it on the downstream task.
This approach (called WikiSub) provides a size-adjusted comparision between using separate upstream data vs the downstream data for pretraining. We see that self-pretraining performs better than WikiSub in the majority of cases (Table~\ref{tab:self_pretrain}).

We also evaluated the alternate pretraining technique \textit{TAPT} 
as described in \citet{gururangan2020don}. 
In this technique, we take the off-the-shelf ELECTRA model,
which has already been pretrained on the upstream BookWiki corpus, 
and further pretrain it on the downstream dataset for 100 epochs.
Self-pretraining outperforms TAPT on 6 datasets,
notably including the two datasets where it outperformed the off-the-shelf models 
by the greatest benefit margin - \textit{AGNews} and \textit{Yahoo Answertopics}.
Interestingly, TAPT performs worse than off-the-shelf model 
on the same 3 datasets where self-pretraining 
performs worse than off-the-shelf model (except Sentiment140).
None of the three pretraining approaches 
seem to be uniformly better than any other.

Finally, we also evaluate the self-pretrained models on the GLUE benchmark and report results on the dev set~\footnote{Following ~\citet{clark2019electra} we exclude the WNLI task from the results.}.
The performance of the models on their pretraining dataset 
does not correlate strongly with its GLUE score.
The GLUE score also does not monotonically go up 
with increasing dataset size, indicating 
that the data domain makes some difference. 
For example, the Amazon Polarity corpus scores just $66.14$ on GLUE 
despite being about 1.4GB in size, 
while AGNews which is 27MB in size, scores $74.30$. 
The highest GLUE score is achieved 
by pretraining on Yahoo Answertopics.

\begin{table*}[h!]
\footnotesize
    \centering
    \setlength{\tabcolsep}{5.0pt}
    \renewcommand{\arraystretch}{1.2}
\begin{tabular}{llrrrrrrl}
\toprule
\textbf{Dataset} &  \textbf{Size(MB)}         &  \textbf{RandInit} &  \textbf{SelfPretrain} &  \textbf{Offshelf} &   \textbf{Benefit\%}  & \textbf{WikiSub}           & \textbf{TAPT}      &   \textbf{GLUE} \\
\midrule
AGNews              &    27 &       91.75 &            94.34 &     93.72 &     131.33 &   93.51   &    94.07 &  74.30 \\
QQP                 &    43 &       82.93 &            90.66 &     90.34 &     104.34 &   89.16   &    90.64 &  75.43 \\
Jigsaw Toxicity     &    59 &       97.83 &            98.49 &     98.53 &      94.99 &   98.35   &    98.48 &  76.65 \\
MNLI                &    65 &       65.49 &            78.39 &     82.29 &      76.77 &   78.64   &    79.26 &  78.28 \\
Sentiment140        &   114 &       63.75 &            67.04 &     66.95 &     102.91 &   65.52   &    65.65 &  72.67 \\
PAWS                &   139 &       50.00 &            97.53 &     97.30 &     100.49 &   97.42   &    97.85 &  74.65 \\
DBPedia14           &   151 &       98.59 &            99.22 &     99.11 &     121.17 &   99.18   &    99.23 &  70.38 \\
Discovery           &   293 &       17.00 &            22.38 &     24.55 &      71.22 &   22.47   &    23.58 &  77.26 \\
Yahoo Answertopics  &   461 &       61.94 &            65.26 &     64.55 &     127.31 &   64.37   &    65.05 &  79.53 \\
Amazon Polarity     &  1427 &       93.86 &            96.27 &     96.13 &     106.49 &   95.82   &    96.16 &  66.14 \\
\bottomrule
\end{tabular}
    \caption{Performance of ELECTRA-small models pretrained with different techniques on various downstream datasets. We also report their performance on the GLUE benchmark (dev set). For reference, a randomly initialized model scores 53.20 and the off-the-shelf model scores 79.43 on GLUE.}
    \label{tab:self_pretrain}
\end{table*}

%% file: sections/50_crosstask_perf.tex
\section{Cross Dataset Finetuning} \label{sec:cross_finetuning}

\begin{figure*}[t!]
    \centering
    \vspace{-5pt}
    \includegraphics[width=0.9\linewidth]{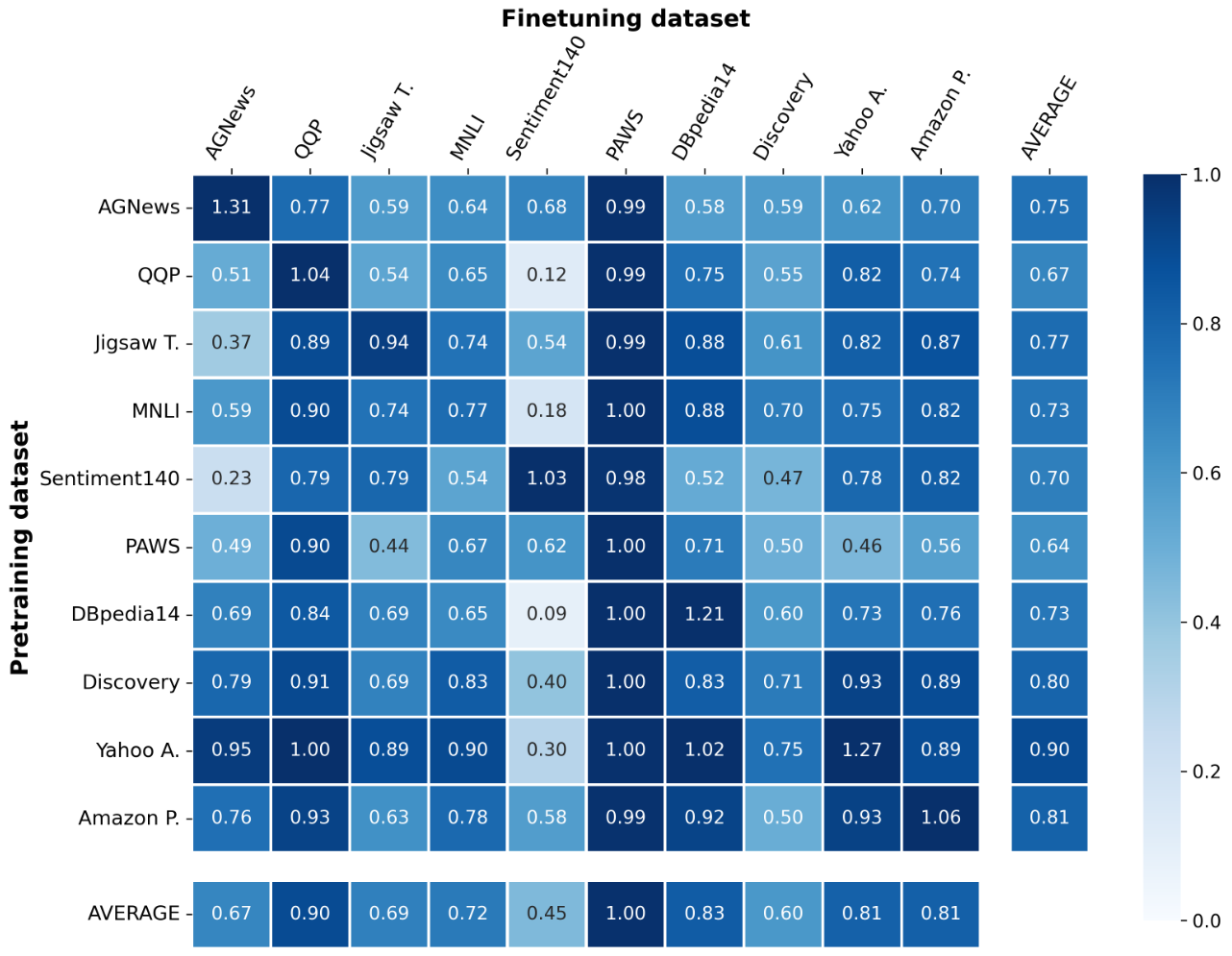}
    \caption{Performance benefits of models pretrained on each dataset, upon finetuning on each downstream dataset. Each value is the ratio of performance gains achieved by model pretrained on the row's dataset vs off-the-shelf model, relative to random initialization, upon finetuning on the column's dataset.}
    \label{fig:cross_finetune_hmap}
\end{figure*}

In this set of experiments, we investigated if the models pretrained on a dataset are only useful for that specific task, or are they useful across the whole spectrum of tasks that we consider.
We took each model pretrained on a dataset in our testbed and finetuned and evaluated it on all other datasets in the testbed. The performance benefits provided in all cases are shown as a heatmap in Figure~\ref{fig:cross_finetune_hmap}.

We found that for almost all downstream datasets, pretraining on any other dataset provides significant advantage (Figure~\ref{fig:cross_finetune_hmap}). 
In most cases, pretraining on the downstream dataset itself performs the best.
Among datasets where self-pretraining performs better than off-the-shelf model (i.e. the diagonal entry is greater than 1), pretraining on datasets of larger size does not help further.
However, for the datasets where self-pretraining's benefit is much less than $100\%$ (i.e. MNLI and Discovery), pretraining on a larger dataset (e.g., Yahoo Answertopics)
performs better than self-pretraining.

Among all the pretrained models, a few models perform consistently good or bad across different downstream datasets (Figure~\ref{fig:cross_finetune_hmap}). 
For example, the model pretrained on Yahoo Answertopics gets the highest average score of 0.90 across all datasets, while the PAWS-pretrained model gives the lowest aggregate score of 0.64. 
Similarly, there are downstream datasets that are benefited consistently by either a large or a small margin by pretraining on different datasets. 
For example, performance on QQP and PAWS receives huge boosts by pretraining on most datasets. 
In contrast, performance on sentiment140 is mostly low , even dropping below 20\% for 3 pretrained models.

We perform an ablation to investigate that given a fixed dataset to finetune on, is it better to pretrain on the \emph{exact} same data (i.e., using the same set of inputs), or is it better to pretrain on different data with an identical distribution.
To test this hypothesis, we divided the training splits of the downstream datasets randomly into two equal subsets (denoted as A and B). We pretrained one model on each subset and then finetuned them on both subsets separately.
The validation and test sets used for finetuning are the same as in the original dataset.

We do not see any consistent benefits with pretraining and finetuning on the same dataset (Table~\ref{tab:split_dataset}). 
Instead, we found consistent patterns where models pretrained on 
one split (either A or B) outperformed models pretrained on the other, irrespective of the split used for finetuning. 
This suggests that the pretraining data has greater influence on the final performance than the finetuning data.
Additionally, we observe that finetuning the superior pretrained model, using the downstream split other than the one used for pretraining, performs the best, 
suggesting overall exposure to more data helps.

\begin{table*}[]
    \centering
    \footnotesize
    \renewcommand{\arraystretch}{1.2}
    \begin{tabular}{rrrrrrrrrrrrrrr}
    
    \toprule
     \multicolumn{3}{c}{\textbf{MNLI}} & & \multicolumn{3}{c}{\textbf{QQP}} & & \multicolumn{3}{c}{\textbf{Discovery}}  & & \multicolumn{3}{c}{\textbf{Yahoo Answertopics}} \\
    \midrule
          & A & B & & &A & B & & &A & B & & &A & B  \\
        A & 76.00 & \textbf{76.42} & & A & 84.28 & 84.79 & & A & 18.78 & 18.61 & & A & 64.18 & \textbf{64.34}  \\
        B & 75.93 & 75.05 & & B & \textbf{88.73} & 88.41 & &  B & \textbf{19.99} & 19.98 & & B & 64.09 & 64.18  \\
    \hline
    \end{tabular}
    \caption{Performance when splitting the dataset into two equal-sized subsets A and B and then pretraining on one (row) and finetuning on another (column)}
    \label{tab:split_dataset}
\end{table*}

%% file: sections/60_ensembling.tex
\section{Difference in Outputs of Self-pretrained and Off-the-shelf Models} \label{sec:disagree}

Since self-pretrained models and off-the-shelf models
perform similarly in terms of classification accuracy,
a natural question to ask is: 
\emph{do these models make errors on the same set of inputs?} 
To answer this question, we investigate the difference in 
predictions made by models pretrained with
different strategies across all multi-class classification tasks. 
In particular, given model $f_A$ and $f_B$, 
we compute \emph{error inconsistency},
defined as follows: 
\begin{align*}
 \sum_{i=1}^n &\frac{\mathbbm{1}\left[ f_A(x_i) \ne y_i \wedge f_B(x_i) = y_i \right]}{n} \\  & \qquad \qquad + \frac{\mathbbm{1}\left[ f_A(x_i) = y_i \wedge f_B(x_i) \ne y_i \right]}{n} \,,   
\end{align*}
where $\{x_i, y_i\}_{i=1}^n$ is the test set. 
Intuitively, error inconsistency captures 
the fraction of examples where exactly one model is correct.
This definition has been commonly used to estimate diversity 
in model prediction~\citep{gontijo2021no,geirhos2020beyond}.  
Across all the multi-class classification tasks, 
in addition to computing error inconsistency 
between self-pretrained and off-the-shelf model, 
for baseline comparison, we also tabulate error inconsistency between: 
(i) two independently finetuned versions of a self-pretrained model; 
and (ii) two independently finetuned versions of the off-the-shelf model.  

Compared to error inconsistency between two 
models with the same pretraining dataset,
we observe that models trained with different pretraining 
datasets have high error inconsistency in predictions (Table~\ref{tab:ensemble}).
For models with comparative performance, 
high error inconsistency highlights 
the high disagreement in predictions. 
This demonstrates that while different pretraining datasets 
produce similarly performing models in terms of overall accuracy, 
the model predictions are relatively dissimilar. 
Our observations here align with investigations in vision tasks, 
where \citet{gontijo2021no} observed that models 
trained with different pretraining datasets 
produced uncorrelated errors. 

Since different pretraining datasets produce models 
with uncorrelated errors, we ensemble 
these models to check if uncorrelated mistakes lead to a correct prediction.
When the models make different predictions, 
in particular,
when one model is correct and another is incorrect, 
the ensemble prediction will be dominated by the model
with higher confidence in their prediction. 
As before, we consider ensembles of 
(i) two independently finetuned self-pretrained models; 
(ii) two independently finetuned off-the-shelf models; 
and (iii) a finetuned version, each of the self-pretrained 
and off-the-shelf models.

We make the following observations:
First, as expected we observe 
that ensembling improves model performance 
as compared to a single model (Table~\ref{tab:ensemble}).
Second, despite having larger error inconsistency,
we do not observe any significant improvements in ensembles of
self-pretrained and off-the-shelf model 
as compared to ensembles of two models with the same pretraining strategy (Table~\ref{tab:ensemble}). 
This is in contrast with findings on vision tasks
where \citet{gontijo2021no} observed 
that larger error inconsistency led to
larger improvement in ensemble performance.

\begin{table*}[t!]
    \centering
    \footnotesize
    \setlength{\tabcolsep}{1.0pt}
    \renewcommand{\arraystretch}{1.2}
\begin{tabular}{lp{0.5cm}cccccp{0.5cm}ccc}
\toprule
 && \multicolumn{3}{c}{\textbf{Ensemble Accuracy}} &&&& \multicolumn{3}{c}{\textbf{Error Inconsistency}} \\
\multirow{2}{*}{  \parbox{1.5cm}{Dataset}} &&  \multirow{2}{*}{ \parbox{2.0cm}{\centering 2$\times$SelfPretrain}} & \multirow{2}{*}{  \parbox{2.0cm}{\centering 2$\times$Offshelf}} & \multirow{2}{*}{  \parbox{1.5cm}{\centering SelfPretrain + Offshelf}} 
&&&& \multirow{2}{*}{  \parbox{2.0cm}{\centering 2$\times$SelfPretrain}} & \multirow{2}{*}{  \parbox{2.0cm}{\centering 2$\times$Offshelf}} & \multirow{2}{*}{  \parbox{1.5cm}{\centering SelfPretrain + Offshelf}}  \\
& & & \\
\midrule
AGNews && $94.66$ & $94.17$ & $94.54$ 
&&&& $1.76$ & $3.50$ & $4.01$ \\
QQP && $90.92$ & $90.74$ & $91.63$ 
&&&& $4.57$ & $5.27$ & $8.91$ \\
MNLI && $78.51$ & $82.37$ & $82.31$ 
&&&& $6.94$ & $6.42$ & $14.82$ \\
PAWS && $97.70$ & $97.45$ & $97.75$ 
&&&& $0.96$ & $1.30$ & $2.07$ \\
DBPedia14 && $99.28$ & $99.19$ & $99.24$ 
&&&& $0.38$ & $0.48$ & $0.51$ \\
Discovery &&  $22.98$ & $25.25$ & $25.02$ 
&&&& $7.85$ & $9.18$ & $12.66$ \\
Yahoo && $65.32$ & $64.69$ & $65.64$ 
&&&& $5.27$ & $5.49$ & $9.55$ \\
Amazon && $96.40$ & $96.24$ & $96.51$ 
&&&& $1.26$ & $1.58$ & $2.48$  \\
\bottomrule
\end{tabular}
    \caption{Performance of ensemble models of self-pretrained and off-the-shelf models. For ensembling, we aggregate predictions of models after calibration with Temperature Scaling~\citep{guo2017calibration}. 
    We observe that in most of the datasets, SelfPretrain + Off-the-shelf ensembling does not improve over ensembles of two models with the same pre-training strategy, despite relatively higher error inconsistency of SelfPretrain + Off-the-shelf models. 
    }
    \label{tab:ensemble}
\end{table*}

%% file: sections/80_generality.tex
\section{Ablations with Other Pretraining Architectures}

We conducted our experiments so far with ELECTRA-small architecture because it is faster to pretrain than other popular models, yet delivers good downstream performance~\citep{clark2019electra}
(e.g. comparable to BERT-base on GLUE benchmark). 
Here, we conduct experiments with a larger model and a different pretraining objective to test the
efficacy of self-pretraining more broadly. 

We experiment with the RoBERTa model 
which uses 
the masked language modeling objective, 
rather than ELECTRA's objective. 
We use the RoBERTa-base architecture,
which has a much larger parameter count of 110 million, 
compared to ELECTRA-small's 14 million. 
Due to resource constraints, we pretrained the RoBERTa models 
for fewer iterations as in ~\citet{warstadt2020learning}. 
We pretrain a RoBERTa-base model 
on the BookWiki corpus for the same number of iterations. 
Our results show that self-pretraining 
performs comparably to pretraining on BookWiki corpus, 
delivering over 85\% of pretraining benefit 
on 9 out of 10 datasets,
and outperforming the model 
pretrained on BookWiki corpus (Table~\ref{tab:roberta_self_pretrain})
on 5 datasets.

\begin{table*}[h!]
    \centering
    \footnotesize
    \setlength{\tabcolsep}{7.0pt}
    \renewcommand{\arraystretch}{1.2}
\begin{tabular}{lrrrrr}
\toprule
\textbf{Dataset} &  \textbf{RandInit} &  \textbf{SelfPretrain} &  \textbf{BookWiki} &   \textbf{Benefit \%} & \textbf{TAPT}   \\
\midrule
AGNews             &       91.91 &            94.28 &          94.22 &     102.27 	&     94.07    \\
QQP                &       76.50 &            88.68 &          90.18 &      89.05 	&     90.64    \\
Jigsaw Toxicity    &       97.32 &            97.72 &          98.03 &      56.02 	&     98.48    \\
MNLI               &       31.82 &            75.12 &          80.90 &      88.23 	&     79.26    \\
Sentiment140       &       56.68 &            68.55 &          60.19 &     338.26 	&     65.65    \\
PAWS               &       50.00 &            97.34 &          97.08 &     100.55 	&     97.85    \\
DBPedia14          &       98.57 &            99.21 &          99.24 &      95.98 	&     99.23    \\
Discovery          &       17.36 &            25.85 &          26.30 &      94.91 	&     23.58    \\
Yahoo Answertopics &       61.11 &            65.96 &          64.58 &     139.80 	&     65.05    \\
Amazon Polarity    &       89.02 &            96.68 &          96.11 &     108.13 	&     96.16    \\
\bottomrule
\end{tabular}
    \caption{Performance of RoBERTa-base models pretrained with different techniques on downstream datasets.}
    \label{tab:roberta_self_pretrain}
\end{table*}

%% file: sections/85_moretasks.tex
\section{Performance on Structured Prediction and Commonsense NLI} \label{sec:SP_task}

While the bulk of our experiments were on a variety of classification tasks, we also experiment with some tasks beyond simple classification. We experiment with three types of tasks: (i) span based question answering, (ii) named entity recognition (NER), and (iii) grounded commonsense inference. For question answering we use the SQuAD dataset~\citep{rajpurkar2016squad} (v1.1) and report the F1-score. For NER, we use the CONLL-2012 NER task which uses annotations from Ontonotes v5.0~\citep{weischedel2013ontonotes} involving 18 kinds of named entities. To measure performance, we use the overall F1 score. We use the seqeval library for evaluation (\url{https://github.com/chakki-works/seqeval}). 
We include SWAG~\citep{zellers2018swag} and HellaSwag~\citep{zellers2019hellaswag} for multiple-choice sentence completion.

For Electra-small models, we see that for each of these datasets self-pretraining achieves more than 70\% pretraining benefit, and for RoBERTa-base model the benefit is 40-80\% (Table~6). Even for the SWAG and HellaSwag datasets, which are designed to use rely on \emph{commonsense inference} of pretrained models, we see performance boosts by pretraining using only the task's training set.

\begin{table*}[h!]
\footnotesize
    \centering
    \setlength{\tabcolsep}{5.0pt}
    \renewcommand{\arraystretch}{1.2}
\begin{tabular}{llrrrrrrrrrr}
\toprule
\textbf{Datasets} &  \textbf{Size(MB)}         &  \multicolumn{4}{c}{\textbf{ELECTRA-small}}  &&&  \multicolumn{4}{c}{\textbf{RoBERTa-base}} \\
&& RI & SP & OS & Benefit\% &&& RI & SP & BW & Benefit\%\\
\midrule
SQuAD             &    19  &       15.82 &            63.01 &     75.96 &      78.47   &  &&     14.93 &            67.23 &     81.89 &      78.11 \\
SWAG              &    22  &       27.55 &            60.56 &     73.76 &      71.43   &  &&     27.95 &            45.18 &     70.37 &      40.62 \\
HellaSwag         &    30  &       29.27 &            39.14 &     42.91 &      72.36   &  &&     24.53 &            31.03 &     34.28 &      66.67 \\
CONLL-2012        &    6.4 &       54.49 &            75.66 &     82.65 &      75.18   &  &&     63.65 &            72.64 &     86.25 &      39.78 \\
\bottomrule
\end{tabular}
    \caption{Performance of ELECTRA and RoBERTa models pretrained with different techniques. RI: random initialization, SP: self-pretraining, OS: off-the-shelf; BW: pretrained on BookWiki by us.}
    \label{tab:extratasks_electra_roberta}
\end{table*}

%% file: sections/90_conclusion.tex
\section{Conclusion and Future Work}

In this work, we showed that pretraining models 
only on text from the downstream dataset 
performs comparably to pretraining 
on a huge upstream corpus 
for a wide variety of datasets. 
The errors made by such \emph{self-pretrained} models on the downstream tasks 
are significantly different from the ones made by the \emph{off-the-shelf} models 
pretrained on upstream corpora.
Our results suggest that the importance of learning from
surplus upstream data for improving downstream task performance 
may have been overestimated.
Crucially, our experiments also do not show that upstream data does not help at all or knowledge transfer does not occur, but simply questions to what extent it is responsible for downstream gains. 
For example, the impressive zero-shot performance very large language models such as GPT-3~\citep{brown2020language} clearly suggests knowledge transfer is involved.
One direction of future work would be to investigate how the performance of self-pretraining compares of pretraining on upstream corpora as the model sizes go up by orders of magnitude.

We found that the quantity and quality of data 
required for pretraining to provide significant benefit 
(over a randomly initialized model trained only with a supervised loss)
is quite low.
Downstream datasets which are tiny in comparison to typical upstream corpora, 
still function as useful pretraining corpora 
for getting performance gains across a wide range of datasets.

Since self-pretraining does not involve any upstream corpus,
it prevents exposure of the model to potentially undesirable contents 
in the large upstream corpus,
while still delivering large performance benefits.
Research has demonstrated the negative influence
of web-sourced pretraining corpora on models, 
such as generating toxic language~\citep{gehman2020realtoxicityprompts} 
or reflecting racial biases in predictions~\citep{ahn2021mitigating}.
For use cases that require avoding such issues, 
self-pretraning can provide a viable alternative to
standard pretraining.
In future work, we hope to compare 
how self-pretrained models and off-the-shelf models 
perform on these negative measures 
such as toxicity and social biases.

%% file: sections/95_limitations.tex
\section{Limitations}

Due to limited availability of compute resources, we were unable to scale up the model architecture to the large sizes becoming increasingly mainstream today. Similarly, the upstream corpus we used (BookWiki) is 16GB in size, and while it is large enough such that it was used to pretrain BERT~\citep{devlin2019bert}, much larger pretraining datasets are in use today such as the Colossal Common Crawl Corpus~\citep{raffel2020exploring}. The relative performance achieved by using self-pretraining vs pretraining on upstream corpus can likely vary with the size of the model and upstream corpus, and more compute-heavy large scale experiments are needed to characterize it.

%% file: sections/appendix.tex
\subsection{The Role of Sentence Order in Pretraining Corpora}

For virtually all pretrained models like BERT, ELECTRA, XLNet, the sentences in the pretraining corpora are ordered as they naturally occur in some document such as Wikipedia article.
\citet{devlin2019bert} mention in their work~:
\emph{``It is critical to use a document-level corpus
rather than a shuffled sentence-level corpus (...) 
in order to extract long contiguous sequences.''}
However, for many of our pretraining corpora made from downstream datasets, the sentence taken in order do not form a coherent document or narrative text.
For example, in the MNLI or QQP corpora, 
neighboring sentences 
will simply be premise-hypothesis pairs 
or potential paraphrase candidates.

Despite the sentence order not forming 
a coherent document, 
many pretraining corpora achieve high performance boosts 
on the GLUE language understanding benchmark (Table~\ref{tab:tfidf_glue}).
For example, MNLI achieves around $96\%$ of the performance boost of the off-the-shelf model (Table~\ref{tab:tfidf_glue}).
Interestingly, shuffling the sentences in these corpora
leads to a large drop in performance (Table~\ref{tab:tfidf_glue}).
This suggests that there is some value to keeping the sentence order in a way that puts sentences from the same example in datasets like MNLI and QQP next to each other.
A likely explanation of this is in \citet{levine2021inductive} where authors showed that including similar sentences in the same input sequence when pretraining should lead to improved performance via theoretical analysis and empirical experiments.

We test if GLUE performance can be improved by artificially re-ordering a set of sentences to promote the occurrence of similar sentences together.
We rearrange the sentences in the sentence-shuffled versions of pretraining corpora to encourage content overlap among neighboring sentences, and see if this can recover some of the drops in performance that occurred due to shuffling. 
Our algorithm creates the corpus by iteratively appending sentences to it, such that at each step the new sentence is the one with maximum TF-IDF similarity with the previous sentence.
Such a way of constructing a corpus by similarity based retrieval has been used in past works~\citep{levine2021inductive,yao2022nlp}, with the main difference that they retrieved sentences from external corpora similar to the ones present in the downstream dataset, whereas we simply use it to reorder sentences already present in the downstream dataset for pretraining
We also make sure that the algorithm does not accidentally recover the original order of sentences (e.g. by matching the premise-hypothesis pairs originally in the MNLI dataset).

We experiment with 5 different datasets and find that the sentence-reordering scheme improves performance compared to random sentence order for all of them except QQP. 
For Discovery and DBPedia14 datasets, it scores even higher than our \emph{standard} sentence ordering scheme which preserves the adjacency and order of sentences within each datapoint. 
This shows that 
re-ordering sentences to promote content similarity between neighboring sentences, can potentially improve GLUE score, without introducing any new information or narrative structure.

\begin{table*}[b]
    \centering
\begin{tabular}{lccc}
\toprule
\textbf{Pretraining Dataset} &  \textbf{Random} &  \textbf{Standard} &  \textbf{TF-IDF(Ours)} \\
\midrule
None (RandomInit)  &     -   &   53.20 &    -   \\
\midrule
Sentiment140 &     -   &   72.67 &  75.29 \\
DBpedia14    &   72.82 &   70.38 &  75.44 \\
Discovery    &   71.79 &   77.26 &  78.94 \\
MNLI         &   62.80 &   78.28 &  76.33 \\
QQP          &   71.09 &   75.43 &  69.57 \\
\midrule
BookWiki (Off-the-shelf)     &     -   &   79.43 &    -   \\
\bottomrule
\end{tabular}
    \caption{GLUE scores achieved by different strategies 
    for ordering sentences from the downstream dataset used for pretraining. 
    Random: randomly ordered sentences; 
    Standard: sentences within a datapoint 
    occur contiguously in original order; 
    TF-IDF: sentences reordered using  content similarity.}
    \label{tab:tfidf_glue}
\end{table*}

\subsection{Experiments with Smaller ELECTRA Models}
\label{sec:tinyelectra}

\begin{figure}[h!]
\centering
\subfigure{\label{fig:a}\includegraphics[width=0.48\textwidth]{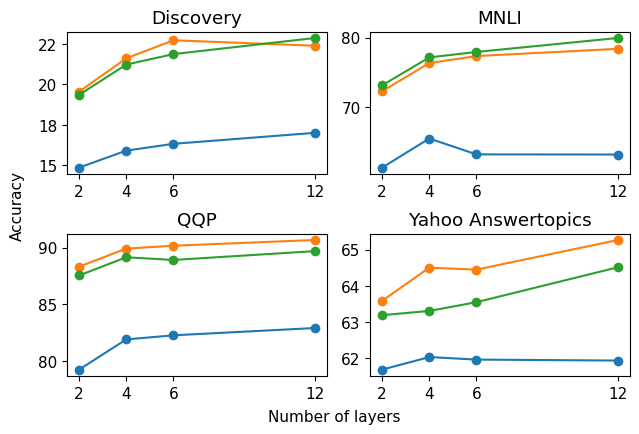}}
\hfill
\subfigure{\label{fig:b}\includegraphics[width=0.48\textwidth]{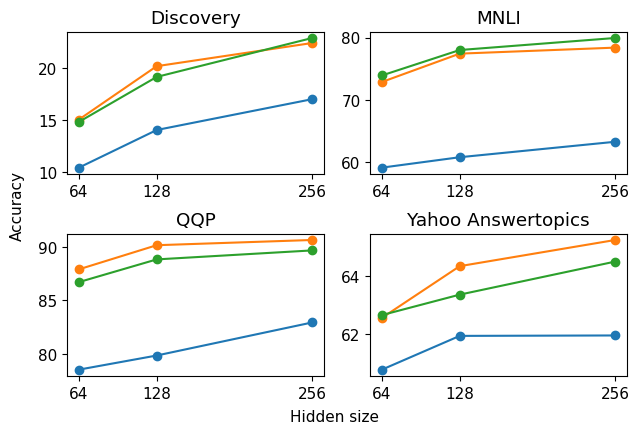}}
\caption{Variation in performance of ELECTRA models with change in number of layers and hidden size 
(\textcolor{tab_blue}{\textbf{---}}~randomly initialized,
\textcolor{tab_orange}{\textbf{---}}~self-pretrained,
\textcolor{tab_green}{\textbf{---}}~BookWiki-pretrained)}
\label{fig:micro_models}
\end{figure}

In addition to experimenting with a \emph{base}-sized architecture (110M parameters), we also experiment with architectures which are even smaller than ELECTRA-small. We train ELECTRA models of smaller size by either reducing the number of layers in the generator and discriminator, or reducing the hidden dimension of the discriminator\footnote{In ELECTRA, the generator's hidden size is already much smaller than that of the discriminator by design. So we do not reduce it further, in order to have a reasonably well-performing generator.}. 
As the models get smaller, self-pretraining continues to significantly outperform random initialization and often outperforms pretraining on BookWiki corpus (Figure~\ref{fig:micro_models}). 
Interestingly, the relative performance of self-pretrained and BookWiki-pretrained models tends to stay the same across model size. For example, for QQP self-pretraining is always best and for MNLI BookWiki-pretraining is always best irrespective of number of layers or hidden size.

\subsection{Implementation Details for Pretraining and Finetuning}

\noindent \textbf{Hyperparameters for Pretraining {} {}}
For pretraining ELECTRA-small models, we use the standard hyperparameters (Table~\ref{tab:hparams_pretrain}) as described in ~\citet{clark2019electra}. For the Roberta-base models, training with the standard hyperparameters with our computing resources would be prohibitively slow, and so we used hyperparameters from ~\citet{warstadt2020learning} which require lesser time to train (Table~\ref{tab:hparams_pretrain}).
For task-adaptive pretraining(TAPT), we follow \citet{gururangan2020don} and further pretrain off-the-shelf models for 100 epochs on the downstream task's training set, with the first 6\% of the resulting total updates used for learning rate warmup.

\noindent \textbf{Hyperparameters for Finetuning {} {}}
For finetuning the models on the 10 downstream datasets,  we use hyperparameters as shown in Table~\ref{tab:hparams_finetune}. We use the AdamW optimizer~\citep{loshchilov2018decoupled} for finetuning. We use early stopping based on  validation set performance. The validation metric used is mean squared error for the sentiment140 dataset (regression), average binary crossentropy for the jigsaw dataset (multi-label classification), and accuracy for all other datasets (multi-class classification). The patience parameter for early stopping is set to 3 epochs.
For finetuning ELECTRA-small models on the GLUE datasets, we use the standard learning rate of 1e-4 following \citet{clark2019electra}.

\noindent \textbf{Details about Use of Downstream Datasets {} {}}
All downstream datasets used in this paper were sourced from the Huggingface library\footnote{\url{https://huggingface.co/docs/datasets/index}}.
For the Yahoo Answertopics dataset, we use only the text from the answer (not the question) as input to the models (both for pretraining and finetuning).
For the PAWS dataset, we use the version called ``Unlabeled $\text{PAWS}_{\text{wiki}}$'' in ~\citet{zhang2019paws}, which is actually \emph{not} unlabeled but has silver labels. We preferred that version over others because of its larger size.
For datasets which had a train and test split but no validation split (e.g. Yahoo Answertopics), we extracted 5000 random datapoints from the the train split to make the validation split.
If a dataset had a train and validation split but no test split (e.g. Unlabeled $\text{PAWS}_{\text{wiki}}$), we designated the validation split to be the test split, and created a new validation set by extracting 5000 random datapoints from the train set.

\begin{table*}[h!]
    \centering
\begin{tabular}{lcc}
\toprule
\textbf{Hyperparameter} & \textbf{ELECTRA} & \textbf{Roberta} \\
\midrule
Size (Parameter count) & Small (14M)  &  Base (110M) \\
Training steps     &  1M     &  100K \\
Warmup steps       &  10K    &   6K \\
Batch size         &  128    &  512 \\
Peak learning rate &  5e-4 &   5e-4 \\
Sequence length    &   128   &   512 \\
\bottomrule
\end{tabular}
    \caption{Hyperparameters used for pretraining models}
    \label{tab:hparams_pretrain}
\end{table*}

\begin{table*}[h!]
    \centering
\begin{tabular}{lcc}
\toprule
\textbf{Hyperparameter} & \textbf{ELECTRA} & \textbf{Roberta} \\
\midrule
Training epochs         &  20     &  20 \\
Batch size              &  32    &  32 \\
Learning rate           &  \{1e-4,1e-5\} &   2e-5 \\
Max sequence length     &   512   &   512 \\
\bottomrule
\end{tabular}
    \caption{Hyperparameters used for finetuning models on 10 downstream tasks}
    \label{tab:hparams_finetune}
\end{table*}

\subsection{Hardware and Software Packages Used}

For pretraining ELECTRA models, we used Nvidia's implementation of the ELECTRA codebase
\footnote{\url{https://github.com/NVIDIA/DeepLearningExamples/tree/master/TensorFlow2/LanguageModeling/ELECTRA}}, run using Nvidia's Tensorflow cotainer image 21.07
\footnote{\url{https://docs.nvidia.com/deeplearning/frameworks/tensorflow-release-notes/rel_21-07.html}}.
For pretraining Roberta models, we used the official implementation in the Fairseq library\footnote{\url{https://github.com/facebookresearch/fairseq}}.
For finetuning experiments, we used the AllenNLP library for training and evaluation routines, coupled with the Huggingface library for the model architectures.

We used a collection of Nvidia V100 (32GB) and A6000(48GB) GPUs for our experiments. Pretraining an ELECTRA-small model takes around 1.5 days on 2 GPUs while pretraining a Roberta-base model takes around 1.5 days on 4 GPUs.